\documentclass[11pt, a4paper,  notitlepage]{extarticle}

\usepackage[left=2.5cm, right=2.5cm, top=3cm, bottom=3cm, bindingoffset=0cm]{geometry}
\usepackage{comment}
\usepackage{graphicx}
\usepackage{mathtools}
\usepackage[affil-it]{authblk}
\usepackage{url}
\usepackage{amsmath}
\usepackage{amssymb}
\usepackage{multicol}
\usepackage{float}
\usepackage[section]{placeins}
\usepackage{hyperref}
\usepackage[title]{appendix}

\usepackage[utf8]{inputenc}
\usepackage{amsmath,amsfonts,amssymb}
\usepackage{graphicx}
\usepackage{url}
\usepackage{booktabs}
\usepackage{array} 

\begin{document}
\sloppy
\title{Multilevel Analysis of Cryptocurrency News using RAG Approach with Fine-Tuned Mistral Large Language Model}
\author{Bohdan M.  Pavlyshenko \\  \small{Ivan Franko National University of Lviv,  Ukraine \\ b.pavlyshenko@gmail.com,  www.linkedin.com/in/bpavlyshenko/ }}
\maketitle

\begin{abstract}
In the paper, we consider multilevel multitask analysis of cryptocurrency news using a fine-tuned Mistral 7B large language model with retrieval-augmented generation (RAG). 
 On the first level of analytics, the fine-tuned model generates graph and text  
summaries with sentiment scores as well as JSON representations of summaries. Higher levels perform hierarchical stacking that consolidates sets of graph-based and text-based summaries as well as summaries of summaries  into comprehensive reports. The combination of graph and text summaries provides complementary views of cryptocurrency news. The model is fine-tuned with 4-bit quantization using the PEFT/LoRA approach. 
The representation of cryptocurrency news as knowledge graph can essentially eliminate problems with large language model hallucinations. 
 The obtained results demonstrate that the use of fine-tuned Mistral 7B LLM  models for  multilevel cryptocurrency news analysis  can conduct informative qualitative and quantitative analytics, providing important insights. 


\end{abstract}

\tableofcontents

\section{Introduction}
Large Language Models (LLMs) with Retrieval-Augmented Generation (RAG) are widely used in text analytics tasks, especially in news analytics. This technology makes it possible to summarize news, extract semantic entities, and generate quantitative scores using few-shot or zero-shot prompts for LLMs. Such an approach is an important component of cryptocurrency analytics. The extracted quantitative scores can then be incorporated into predictive models in subsequent stages of analysis.
Large language models (LLM), based on generative pre-trained transformers (GPT),  such as ChatGPT,  show high efficiency in the analysis of complex texts. 
Nowadays, we can observe the emergence of many new smaller open source LLMs, e.g. Llama, Falcon, GPT4All, GPT-J, Mistral, etc. 
Open source LLMs can be fine-tuned for specific custom problems and deployed on custom servers, e.g. in cloud computing services such as AWS, GCP. 
 LLMs have some new features as compared to conventional language models based on transformers. One of them is zero-shot and few-shot learning, which 
 consists in a good performance of the model when we show only a few training examples to it or even no examples at all, but only the instructions describing what should be done. Another important feature is the reasoning when a model can generate new patterns and conclusions which are based on an input prompt and facts known by the model and which were not included into it directly during a training process. So, the model can generate analytical texts with unexpected but useful chains of thoughts.  

In the articles \cite{pavlyshenko2025ai, pavlyshenko2023financial, pavlyshenko2023analysis}, we consider different approaches of using LLMs in text analytics. 
In predictive analytics, it is important to take into account informational trends from different data sources, including news websites and social networks. 
  The paper~\cite{pavlyshenko2022forming} considers a number of approaches for forming different predictive features of tweet data sets and using them in predictive analysis for the decision-making support. The graph theory as well as frequent itemsets and association rules theory is used for forming and retrieving different features from these datasests. The use of these approaches makes it possible to reveal a semantic structure in tweets related to a specified entity. It is shown that quantitative characteristics of semantic frequent itemsets can be used in predictive regression models with specified target variables. Using  the graph theory, users’ communities and influencers can be revealed given tweets characteristics. 
In \cite{pavlyshenko2022methods},  different approaches for the analysis of news trends on X (formerly Twitter)  have been considered.  
The obtained results show that an effective system for detecting fake and manipulative news can be developed using combined neural network which consists of three concatenated subnetworks. 
LLMs are being effectively used in analysing  financial data and news. The paper~\cite{yang2023fingpt} 
considers an open-source large language model, FinGPT, for the finance sector.  FinGPT takes a data-centric approach, providing researchers and practitioners with accessible and transparent resources to develop their FinLLMs. 
Different  approaches including using LLM for text analytics are considered in~\cite{chen2023_chatgpt_gnn,  ibrahim2024survey, agrawal2023can, 
shang2024_llmgga}.
These approaches demonstrate how GPT-style LLMs can structure financial text into semantic or relational graphs, feeding into GNNs or knowledge-graph pipelines for prediction, risk assessment, recommendation, and fraud detection.
In~\cite{pavlyshenko2023financial}, authors apply PEFT/LoRA to Llama 2 GPT, fine-tuned for multiple tasks: summarization, key-point extraction, and entity extraction with sentiment. The structured outputs (in JSON) allow sentiments tied to named entities---a powerful feature for downstream ML prediction tasks.
In~\cite{efeoglu2024relation}, authors fine-tuned LLMs---Mistral-7B, Llama 2-7B, T5 Large---for relation extraction across datasets. They showed that fine-tuning dramatically improves the extraction of implicit and semantic relationships, particularly with Mistral-7B. 
The fine-tuned Mistral-7B model can be used for detecting and analyzing fake news, propaganda and offensive language in news articles~\cite{bpavlshmistrfnews2025}.  Given the news text, the model detects and analyses fake news and propaganda, analyses and shows manipulative constructions in the text as well as shows offensive language.
The fine-tuned Seq2Seq model is developed for analysing and summarizing cryptocurrency news for the following crypto coins: Bitcoin, Ethereum, Tether, Solana, Binance Coin~\cite{bpavlcryptosumm2024}. The model is created by fine-tuning the facebook/bart-large transformer model. The model outputs short text summary and uptrend/downtrend lists of specified above crypto coins if their trends are considered in the news text.
In~\cite{pavlyshenko2014clustering}, we considered the use of semantic fields theory for text analytics. The paper~\cite{pavlyshenko2019bitcoin} describes the linear model for Bitcoin price which includes regression features based on Bitcoin currency statistics, mining processes, Google search trends and Wikipedia pages visits.  It is shown that Bayesian approach makes it possible to utilize the probabilistic approach using distributions with fat tails and take into account the outliers in Bitcoin price time series.

 One of the approaches of using LLMs is based on retrieval-augmented generation (RAG), which uses the results from other services, e.g. relational database, semantic search, graph database in the input prompt for LLM. In this case,   the response  can be treated as a combination of  external results and LLM knowledge.
Retrieval-Augmented Generation (RAG)  enhance large language models
(LLMs) by integrating external knowledge sources, enabling more accurate and
contextually relevant responses tailored to user needs. 
Different approaches for using RAG are considered in ~\cite{edge2024local,procko2024graph, lewis2020rag, he2024g,peng2024graph,jin2024graph,zhang2025survey, guo2024lightrag, xu2024retrieval}
Retrieval-Augmented Generation (RAG) has achieved remarkable success in addressing the challenges of Large Language Models (LLMs) without necessitating retraining. By referencing an external knowledge base, RAG refines LLM outputs, effectively mitigating issues such as ``hallucination'', lack of domain-specific knowledge, and outdated information. 

In this study, we are going to consider and test the use of the PEFT/LoRA approach for fine-tuning Mistral 7B large language model (LLM) on  cryptocurrency news dataset on multilevel multitask instructions: graph and text summarizing, JSON representation of summaries, and stacking a list of graph and text summaries.   

\section{Graph Representation of Text in News Analytics Using RAG Approach}
Graph representation of data and semantic entities are widely used in text analytics. 
The use of graph approaches in text analytics can give us the following benefits:
\begin{itemize}
\item Structured Knowledge \& Relationship Discovery.
Using large language models (LLMs) to generate graph representations of news, for instance, knowledge graphs or content‐entity graphs enable deep analytics with several key benefits~\cite{ge2020graph,shang2024survey,mao2024advancing}.
\item Enhanced Reasoning \& Fact Validation.
Unlike free text, graphs enable more effective way to reason over news statements (e.g. “X acquired Y on date Z”) and check credibility. When news is represented in a knowledge graph, cross-checking claims become systematic.
\item Better Recommendations \& Trend Analytics.
In news recommendation systems, incorporating graph structures (e.g. user-click history, related article networks) improved representation learning for both articles and users. These graph-enhanced models substantially boost personalization quality. 
\cite{ge2020graph}
\item LLM‑Graph Hybrid Reasoning.
LLMs excel at semantic extraction from texts; graphs encode that information in relational form. Surveys show that combining LLMs with graph neural networks (GNNs) or knowledge representation frameworks enables better link prediction, node classification, and complex reasoning over news data. Compared to standalone LLM or GNN, hybrids generalize more effectively. \cite{jin2024large,wang2025large}
\end{itemize}
In survey~\cite{peng2024graph}, the use of graph retreival-augmented generation has been considered. 
The survey~\cite{zhang2025survey} presents a systematic analysis of Graph-based Retrieval-Augmented Generation (GraphRAG), a new paradigm that revolutionizes domain-specific LLM applications.
The paper \cite{mao2024advancing} reviews methods and architectures combining LLMs with graph representation learning, detailing design strategies and capability gains.
The use of knowledge graphs can reduce hallucinations in LLMs~\cite{agrawal2023can,  lavrinovics2025knowledge, lavrinovics2025knowledge}.
Graph representations - whether used before, during, or after generation — provide the structure, verification, and grounding that effectively mitigate LLM hallucinations. They 
anchor outputs in factual triples, 
enable automated claim for checking and refinement.
 Dynamic knowledge graphs  capture evolving relationships over time, enabling early signal detection.
Extracting sentiment from financial news and embedding it in forecasting models leads to improved prediction accuracy.
 Knowledge-graph structures provide clear lineage of claims or relationships, improving trust, especially in regulated trading environments. 
Traditional RAG systems segment documents into fixed chunks and retrieve relevant passages using vector similarity search. However, this approach suffers from fundamental limitations when addressing queries requiring complex reasoning across interconnected information sources.
 When answers require synthesizing knowledge from multiple, seemingly unrelated sources, vector-based similarity often cannot capture the necessary semantic connections.  Traditional RAG also have problems with  holistic understanding of entire datasets, where global patterns and themes are more important than specific factual details. 
Recent advances in graph-enhanced RAG architectures address these limitations by incorporating structured knowledge representations. GraphRAG \cite{edge2024local} introduces hierarchical community detection to create structured summaries of semantically related entities. HippoRAG \cite{gutierrez2024hipporag} draws inspiration from hippocampal memory processes to enable associative retrieval through knowledge graphs. RAPTOR  \cite{sarthi2024raptor} employs recursive tree-based organization for hierarchical document understanding.
Knowledge graphs provide structured representations of entities and their relationships, enabling sophisticated reasoning capabilities. Graph-based approaches offer several advantages: explicit relationship modeling,  hierarchical organization of concepts, and support for complex query types requiring structural reasoning.
Microsoft Research's GraphRAG \cite{edge2024local} introduces a two-stage indexing approach combining entity knowledge graph extraction with hierarchical community detection. 
Recent advances in Retrieval-Augmented Generation (RAG) have demonstrated the potential of graph-based representations for improving large language model (LLM) performance on complex reasoning tasks. 
Graph-based RAG approaches for news understanding offer significant advantages in terms of accuracy, reasoning depth, and interpretability. With the rising complexity of real-world information ecosystems, integrating LLMs with structured graph reasoning is a promising direction for future research.

\section{Using Stacking Approach in News Analytics}
Stacking approach is widely used in predicitive analytics with numerical target variables for improving the accuracy of forecasting results. In the paper~\cite{pavlyshenko2019machine}, we study the usage of machine-learning models for sales predictive analytics. The main goal of this paper is to consider main approaches and case studies of using machine learning for sales forecasting. A stacking approach for building regression ensemble of single models has been studied. The results show that using stacking techniques, we can improve the performance of predictive models for sales time series forecasting.
In~\cite{pavlyshenko2020using}, we study the Bayesian regression for building time series models and stacking different predictive models for time series. The use of Bayesian regression for time series modeling with nonlinear trend was analyzed. This approach makes it possible to estimate an uncertainty of time series prediction and calculate value at risk characteristics.  The probabilistic approach for stacking predictive models allows us to make risk assessment for the predictions that are important in a decision-making process.
Stacking approach makes it possible to combine LLM models output on the next analytical level. 
It can be considered as a summary of summaries. As a result, one can get diversified optimized results.  
Key Benefits of Stacking GPT Responses:
\begin{itemize}
\item Improved Accuracy \& Robustness via Meta-Learner Integration.
Stacking leverages diverse GPT outputs as “base-level” responses, which are then combined by a second-stage model (a meta‑learner) that learns to weigh or correct them. This reduces individual model bias and captures complementary strengths across models 
\item Consensus Reasoning \& Error Mitigation.
Frameworks like Iterative Consensus Ensemble (ICE) refine outputs through repeated passes, building agreement across GPT outputs. This reduces hallucinations and increases reliability of answers, since each iteration reinforces consistent responses.
\item Hierarchical Summarization for Multi‑Level Insights.
In the news analytics domain, a two-tier pipeline first generates detailed summaries or entity‑relation graphs from individual articles, then synthesizes them into a meta‑summary. 
\item Quantitative Opinion Scoring.
In applications like opinion and trend tracking, the first stage may produce sentiment scores or entity‑relationship extractions per article; the second stage aggregates these into trend scores, uncertainty estimates, or network summaries using Bayesian or graph analytic meta‑models. 
\end{itemize}

\section{Parameter-Efficient Fine-Tuning Mistral 7B LLM}

Full fine-tuning large language models (LLMs) are computationally expensive and memory-intensive. 
It can be applicable in the case when we need to ingest millions of documents into LLM. But in case of much smaller data, we can use a PEFT/LoRA approach which consists in fine-tuning a much smaller number of model parameters. These parameters are saved in the model adapter which is used for full model modification before using it for the model text response generation.  To optimize GPU usage, 
4bit or 8bit quantization of LLM can be chosen for model fine-tuning.
State-of-the-art Parameter-Efficient Fine-Tuning (PEFT) methods enable efficient adaptation of pre-trained language models (PLMs) to various downstream applications without fine-tuning all the model's parameters. The fine-tuning of large-scale PLMs is often prohibitively costly. In this regard, PEFT methods only fine-tune a small number of  model parameters, thereby greatly decreasing the computational and storage costs. Recent State-of-the-Art PEFT techniques achieve the performance comparable to that of full fine-tuning~\cite{peft}.
The paper~\cite{hu2021lora} considers Low-Rank Adaptation, or LoRA, which freezes pre-trained model weights and injects trainable rank decomposition matrices into each layer of the Transformer architecture, greatly reducing the number of trainable parameters for downstream tasks.  PEFT/LoRA approach makes it possible to fine-tune LLMs with sizes near 7B parameters, using Google Colab.  Along with text data for fine-tuning, it is important to use prompt instructions which show how to process input prompts. Instructions can be created by human experts and augmented by other LLM models. 
LLM generate complex output texts on prompts which can be optimized in different ways. One possible way is selecting appropriate instructions for fine-tuning models. Another way is using a method called Reinforcement Learning from Human Feedback (RLHF)~\cite{beeching2023stackllama}. In this approach, human experts estimate and rate LLM output and then using this rates as target lables, LLM can be fine-tuned by supervised training.  
PEFT methods aim to reduce resource demands while maintaining performance.
PEFT/LoRA approaches are described in~\cite{hu2021lora, dettmers2023qlora, dettmers2022bitsandbytes, peft, qloraGithub, bitsandbytesGithub, transformersGithub}. 
Mistral is a family of open-weight decoder-only large language models developed by Mistral AI, designed for high performance, efficient inference, and open availability. The architecture follows a transformer decoder design with key improvements for speed and memory efficiency~\cite{mistralHF,mixtralHF}.
The model has the following main features:
sparse Attention via Sliding Window Attention for better scaling to long sequences;
grouped-Query Attention (GQA) for faster inference compared to Multi-Query Attention;
efficient training using large-scale open corpora and optimized data pipelines.
Instruction-tuned variants released (e.g., Mistral-Instruct, Mixtral) for chatbot and assistant tasks.
Notable Models:
Mistral-7B, 7 billion parameters, trained from scratch, competitive with larger models;
Mixtral-8x7B, a Mixture of Experts (MoE) model using 8 experts (2 active per token), combining high quality and computational efficiency~\cite{mistral7b, mistralHF, mixtralHF}.
Let us consider fine-tuning Mistral 7B using a PEFT/LoRA approach. 
The instruction for fine-tuning can be created in different ways, e.g. by experts or using LLMs  with appropriate prompts. 
 For the model fine-tuning, the trainer \verb|SFTTrainer| 
from package \verb|trl|~\cite{vonwerra2022trl} was used.
We have conducted fine-tuning Mistral 7B model on a mixture of tasks using PEFT/LoRA approach with 4-bit quantization. 
The following training arguments were set up:
\begin{newmargin}{1cm}{0cm} 
\begin{verbatim}
per_device_train_batch_size=2,
gradient_accumulation_steps=4,
per_device_eval_batch_size=2,
eval_strategy="steps",
eval_steps=500,
save_steps=500,
logging_steps=500,
num_train_epochs=7,
save_strategy="steps",
learning_rate=3e-5,
fp16=True
\end{verbatim}
\end{newmargin}

For multilevel analytics, we have considered the following tasks:
\begin{itemize}
\item Creating knowledge graph on cryptocurrency news articles;
\item Summarizing crypto currency news with generating quantitative sentiments scores for cruptocurrencies considered in the article;
\item Generating JSON representation of knowledge graph and text summaries;
\item Summarizing the list of knowledge graphs and the list of text summaries with sentiment scores;
\item Generating stacking summary of the summary of graph summaries list and the summary of text summaries list.
\end{itemize} 

Using both graph and text summaries allows us to create a diversified representation of crypotcurrency news. Such an approach enables us to receive comprehensive summary on the stacking level where similarity and diversity in cryptocurrency trends can be highlighted. 
Representation cryptocurrency news as knowledge graph can essentially eliminate problems with large language model hallucinations.

\subsection{Prompt for Fine-Tuning Mistral 7B LLM}
For fine-tuning Mistral 7B model, we used the following prompt:
{\fontsize{11}{12}\selectfont \begin{verbatim}
<s>[INST] <<SYS>>
You are an expert in analyzing cryptocurrency news.
<</SYS>>

{Prompt query}
{Text} [/INST] {LLM Response} </s>
\end{verbatim}
}
where \verb|{Prompt query}| is the replacement field for the prompt query of different analytical tasks, 
\verb|{Text}| is the replacement field for a text of financial news, \verb|{LLM Response}| is the replacement field for LLM's analytical response. 
Depending on the \verb|{Prompt query}|, the fine-tuned model generates responses on different analytical levels with the following tasks:
\begin{itemize}
\item On the first level, the model generates two types of summaries of cryptocurrency news: graph summary and text summary. The prompt query on this level: \textit{"Generate a knowledge graph from cryptocurrency news:"} and 
\textit{"Generate summaries of cryptocurrency news and detect sentiment signals:"}.
\item JSON representation of summaries received on the first level.
The prompt query on this level: \textit{"Create a JSON representation of the summary of cryptocurrency news:"}.
\item Summarazing the list of graph or text summaries. Several summaries from the first level can be combined into one prompt, we used 5 arbitrary chosen summaries for one prompt.
The prompt query on this level: \textit{"Summarize the following list of cryptocurrency news summaries:"}. 
On this analytical level, \verb|{Text}| is a list of graph or text summaries of cryptocurrency news in the format:
{\fontsize{11}{12}\selectfont \begin{verbatim}
"""
News:{i}
    
{Graph or text summary}
------------------------------------------------------
"""
\end{verbatim}
}
\break
\item Generating stacking summary of the summary of graph summaries list and the summary of text summaries list. The prompt query on this level: \textit{"Generate a single summary from the two provided summaries and output the result in JSON format:"}. For this task, we used the following format for the replacement field \verb|{Text}|:
{\fontsize{11}{12}\selectfont \begin{verbatim}
"""

Summary 1:
{Stacking graph summary}

Summary 2:
{Stacking text summary}
"""
\end{verbatim}
}
\end{itemize}

For the model training, we prepared the instructions for multilevel analytics which consist of tasks considered above.
Multilevel stacking approaches of news with graph and text summary representation make it possible to receive a diversified summary and highlight strong trends which are similar in most of news and contradictory trends and facts which have different explanation in different news. That is important for the decision-making on the expert level, allows one to make risk assessments and eliminate LLM hallucinations in news summarization.

\subsection{Knowledge Graph Generation}
We used the prompts for GPT-4.1 model to generate training datasets of instructions. For describing a graph structure, we used the following descriptions of graph entities and relationships:
{\fontsize{11}{12}\selectfont \begin{verbatim}
<Cryptocurrency> — has_sentiment_signal — <score>

Entity types to extract:
- Cryptocurrency (e.g., Bitcoin, Ethereum)
- Person (e.g., Elon Musk)
- Organization (e.g., Binance, SEC)
- Exchange (e.g., Coinbase, Kraken)
- Wallet/Provider (e.g., MetaMask, Ledger)
- Event (e.g., ETF approval, hack)
- Regulation or Law
- Country or Government Body
- Project or Protocol (e.g., Uniswap, Solana)
- Metric (e.g., price, volume, market cap)
- Sentiment (bullish, bearish, positive, negative)
- Upward trend
- Downward trend

Relation types to extract:
- launched_by
- regulated_by
- affected_by
- involved_in
- listed_on
- invested_by
- announced
- supported_by
- discussed_in
- target_of
- resulted_in
- has_price_change
- predicted_by
- acquired_by
- has_trend
- has_sentiment_signal (scale from -10 to 10)
\end{verbatim}
}

\section{Testing Fine-Tuned Mistral 7B Model}
For testing, we loaded arbitrary cryptocurency news from the Web for July 2025. 
The fine-tuned model can be tested using the code given in the Appendix. 
PEFT adapter of the fine-tuned model has been placed on HuggingFace platform~\cite{pavlshcrypto2025}.

\subsection{Prompt for Testing Fine-Tuned Mistral 7B Model}
For testing the fine-tuned model, we used the prompt similar to the fine-tuning prompt but without the final part "\verb|{LLM Response} </s>|" which will be generated by the model:
{\fontsize{11}{12}\selectfont \begin{verbatim}
<s>[INST] <<SYS>>
You are an expert in analyzing cryptocurrency news.
<</SYS>>

{Prompt query}
{Text} [/INST]
\end{verbatim}
}
For higher stacking levels, we used a similarly modified prompt created for model fine-tuning on stacking tasks.  

On the stacking level, the model can detect contradictory trends in the list of news summaries. 
As a result of multilevel cryptocurrency news analytics, one can receive a diversified analytical summary with eliminated hallucinations which include diversified information accumulated from the list of cryptocurrency news. 

\subsection{Examples of Cryptocurrency News Analytics Using Fine-Tuned Mistral 7B Model}
Here we give some results of cryptocurrency news processing, others can be found in the Appendix. 
\subsubsection{Example of graph summary}
{\fontsize{9}{11}\selectfont \begin{verbatim}
 Ethereum — has_trend — upward trend
Ethereum — has_sentiment_signal — 8
Ethereum — affected_by — institutional interest
Ethereum — affected_by — stablecoin legislation
Ethereum — affected_by — stablecoin market cap
Ethereum — affected_by — Circle IPO
Ethereum — affected_by — new leadership
Ethereum — affected_by — treasury firms
Ethereum — affected_by — corporate entrants
Ethereum — affected_by — ETH ETFs
Ethereum — affected_by — ether staking
Ethereum — affected_by — Merge transition
Ethereum — affected_by — increased competition from Solana
Ethereum — affected_by — lower transaction fees
Ethereum — affected_by — slower speeds
Ethereum — affected_by — tokenization of financial assets
Ethereum — affected_by — institutional demand
Ethereum — affected_by — decentralization
Ethereum — affected_by — tokenization of dollars
Ethereum — affected_by — institutional arc
Ethereum — has_sentiment_signal — 8
Solana — involved_in — competition with Ethereum
Solana — has_sentiment_signal — 0
Circle — involved_in — IPO
Circle — involved_in — second-largest stablecoin
Circle — involved_in — legislation around stablecoins
Ethereum — has_price_change — $4,000
Bitcoin — compared_to — Ethereum (10 years ago)
Bitcoin — has_sentiment_signal — 0
Stablecoins — launched_on — Ethereum
Stablecoins — has_market_cap — $100 billion
Crypto in general — affected_by — institutional interest
Crypto in general — affected_by — tokenization of financial assets
Crypto in general — affected_by — tokenization of dollars
Crypto in general — affected_by — tokenization of financial assets
Crypto in general — affected_by — institutional demand
Crypto in general — affected_by — tokenization of dollars
Crypto in general — affected_by — tokenization of financial assets
Crypto in general — affected_by — institutional demand
Crypto in general — has_sentiment_signal — 2
\end{verbatim}
}
\subsubsection{Example of graph summary in JSON format}
{\fontsize{9}{11}\selectfont \begin{verbatim}
 [
  {"Entity 1":"Ethereum","Relationship":"has_trend","Entity 2":"upward trend"},
  {"Entity 1":"Ethereum","Relationship":"has_sentiment_signal","Entity 2":"8"},
  {"Entity 1":"Ethereum","Relationship":"affected_by","Entity 2":"institutional interest"},
  {"Entity 1":"Ethereum","Relationship":"affected_by","Entity 2":"stablecoin legislation"},
  {"Entity 1":"Ethereum","Relationship":"affected_by","Entity 2":"stablecoin market cap"},
  {"Entity 1":"Ethereum","Relationship":"affected_by","Entity 2":"Circle IPO"},
  {"Entity 1":"Ethereum","Relationship":"affected_by","Entity 2":"new leadership"},
  {"Entity 1":"Ethereum","Relationship":"affected_by","Entity 2":"treasury firms"},
  {"Entity 1":"Ethereum","Relationship":"affected_by","Entity 2":"corporate entrants"},
  {"Entity 1":"Ethereum","Relationship":"affected_by","Entity 2":"ETH ETFs"},
  {"Entity 1":"Ethereum","Relationship":"affected_by","Entity 2":"ether staking"},
  {"Entity 1":"Ethereum","Relationship":"affected_by","Entity 2":"Merge transition"},
  {"Entity 1":"Ethereum","Relationship":"affected_by","Entity 2":"increased competition from
	Solana"},
  {"Entity 1":"Ethereum","Relationship":"affected_by","Entity 2":"lower transaction fees"},
  {"Entity 1":"Ethereum","Relationship":"affected_by","Entity 2":"slower speeds"},
  {"Entity 1":"Ethereum","Relationship":"affected_by","Entity 2":"tokenization of financial
	assets"},
  {"Entity 1":"Ethereum","Relationship":"affected_by","Entity 2":"institutional demand"},
  {"Entity 1":"Ethereum","Relationship":"affected_by","Entity 2":"decentralization"},
  {"Entity 1":"Ethereum","Relationship":"affected_by","Entity 2":"tokenization of dollars"},
  {"Entity 1":"Ethereum","Relationship":"affected_by","Entity 2":"legislation around stablecoins"},
  {"Entity 1":"Ethereum","Relationship":"has_price_change","Entity 2":"$4,000"},
  {"Entity 1":"Ethereum","Relationship":"affected_by","Entity 2":"10 years ago (tokenization of
	dollars)"},
  {"Entity 1":"Bitcoin","Relationship":"compared_to","Entity 2":"Ethereum (10 years ago)"},
  {"Entity 1":"Bitcoin","Relationship":"has_sentiment_signal","Entity 2":"0"},
  {"Entity 1":"Stablecoins","Relationship":"launched_on","Entity 2":"Ethereum"},
  {"Entity 1":"Stablecoins","Relationship":"has_market_cap","Entity 2":"$100 billion"},
  {"Entity 1":"Stablecoins","Relationship":"affected_by","Entity 2":"legislation around
	stablecoins"},
  {"Entity 1":"Circle","Relationship":"involved_in","Entity 2":"IPO"},
  {"Entity 1":"Circle","Relationship":"involved_in","Entity 2":"second-largest stablecoin"},
  {"Entity 1":"Circle","Relationship":"involved_in","Entity 2":"legislation around stablecoins"},
  {"Entity 1":"Circle","Relationship":"has_sentiment_signal","Entity 2":"8"},
  {"Entity 1":"Ethereum","Relationship":"has_sentiment_signal","Entity 2":"8"},
  {"Entity 1":"Solana","Relationship":"involved_in","Entity 2":"competition with Ethereum"},
  {"Entity 1":"Solana","Relationship":"has_sentiment_signal","Entity 2":"0"},
  {"Entity 1":"Crypto in general","Relationship":"affected_by","Entity 2":"institutional interest"},
  {"Entity 1":"Crypto in general","Relationship":"affected_by","Entity 2":"tokenization of financial
	assets"},
  {"Entity 1":"Crypto in general","Relationship":"affected_by","Entity 2":"institutional demand"},
  {"Entity 1":"Crypto in general","Relationship":"affected_by","Entity 2":"legislation around
	stablecoins"},
  {"Entity 1":"Crypto in general","Relationship":"has_sentiment_signal","Entity 2":"2"}
]
\end{verbatim}
}
\subsubsection{Example of text summary}
{\fontsize{9}{11}\selectfont \begin{verbatim}
 **Summary:**
Ethereum (ETH) is experiencing renewed institutional interest, driven by the U.S. stablecoin law,
new leadership, and increasing recognition of its role in decentralized finance and stablecoin
issuance. Despite past struggles with high costs and competition from Solana, Ethereum's
institutional appeal is growing, though its price remains below its 2021 highs. The current bullish
sentiment is supported by new treasury firms, ETF inflows, and a focus on Ethereum's
decentralization.


**Crypto Descriptions and Sentiment Scores:**

- **Ethereum (ETH):**
  Ethereum is a decentralized blockchain platform designed to support complex applications, and is
currently benefiting from institutional adoption, new treasury firms, and regulatory clarity,
despite earlier struggles with high costs and competition.
  **Sentiment Signal Score:** +8

- **Stablecoins (Crypto in general):**
  Stablecoins are cryptocurrencies pegged to stable assets like the U.S. dollar, and their
increasing use in cross-border transactions is driving institutional interest in blockchain
technology.
  **Sentiment Signal Score:** +7

- **Bitcoin (BTC):**
  Bitcoin is the original decentralized cryptocurrency, and its success in 2021 helped establish
Ethereum's potential as a more versatile network.
  **Sentiment Signal Score:** 0

- **Solana (SOL):**
  Solana is a high-speed blockchain competing with Ethereum for network capacity and transaction
fees, but Ethereum's institutional appeal remains stronger.
  **Sentiment Signal Score:** -2

- **Circle (CRCL):**
  Circle is the issuer of the second-largest stablecoin, USDC, and its recent acquisition by French
media company Meteor signifies further institutionalization of stablecoin issuance.
  **Sentiment Signal Score:** +6

- **Crypto in general:**
  Cryptocurrencies are increasingly being adopted by institutions for decentralized finance,
tokenization of traditional assets, and regulatory compliance, with broader mainstream acceptance
accelerating.
  **Sentiment Signal Score:** +7
\end{verbatim}
}
\subsubsection{Example of text summary in JSON format}
{\fontsize{9}{11}\selectfont \begin{verbatim}
 [
  {
    "Cryptocurrency": "Ethereum (ETH)",
    "summary": "Ethereum is a decentralized blockchain platform designed to support complex
applications, and is currently benefiting from institutional adoption, new treasury firms, and
regulatory clarity, despite earlier struggles with high costs and competition.",
    "sentiment": 8
  },
  {
    "Cryptocurrency": "Stablecoins (Crypto in general)",
    "summary": "Stablecoins are cryptocurrencies pegged to stable assets like the U.S. dollar, and
their increasing use in cross-border transactions is driving institutional interest in blockchain
technology.",
    "sentiment": 7
  },
  {
    "Cryptocurrency": "Bitcoin (BTC)",
    "summary": "Bitcoin is the original decentralized cryptocurrency, and its success in 2021 helped
establish Ethereum's potential as a more versatile network.",
    "sentiment": 0
  },
  {
    "Cryptocurrency": "Solana (SOL)",
    "summary": "Solana is a high-speed blockchain competing with Ethereum for network capacity and
transaction fees, but Ethereum's institutional appeal remains stronger.",
    "sentiment": -2
  },
  {
    "Cryptocurrency": "Circle (CRCL)",
    "summary": "Circle is the issuer of the second-largest stablecoin, USDC, and its recent
acquisition by French media company Meteor signifies further institutionalization of stablecoin
issuance.",
    "sentiment": 6
  },
  {
    "Cryptocurrency": "Crypto in general",
    "summary": "Cryptocurrencies are increasingly being adopted by institutions for decentralized
finance, tokenization of traditional assets, and regulatory compliance, with broader mainstream
acceptance accelerating.",
    "sentiment": 7
  }
]
\end{verbatim}
}

\subsubsection{Example of stacking graph summaries}
{\fontsize{9}{11}\selectfont \begin{verbatim}
 Upward trend: Bitcoin (strong upward trend driven by record-high price peaks, institutional
accumulation, positive sentiment, and significant ETF inflows), Crypto in general (overall upward
trend supported by positive sentiment, institutional activity, and regulatory progress such as the
Genius Act).

Downward trend: None identified with strong downward trend in the provided news.

Contradictory trend: Bitcoin (contradictory signals with both strong upward momentum from
institutional activity and ETF inflows and significant downward price movement and sentiment in the
same news cycle).

Conclusion: Bitcoin is experiencing a strong upward trend due to institutional accumulation and ETF
inflows, but faces contradictory signals from recent price drops and negative sentiment; Crypto in
general is buoyed by positive sentiment, regulatory progress, and institutional activity; no clear
strong downward trend is identified in the provided news.

\end{verbatim}
}

\subsubsection{Example of stacking text summaries}
{\fontsize{9}{11}\selectfont \begin{verbatim}
Upward trend: Bitcoin (BTC) - driven by strong institutional inflows, regulatory clarity from U.S.
lawmakers, and a new all-time high above $123,000; Ethereum (ETH) - supported by growing
institutional adoption, tokenized real-world asset offerings, and positive sentiment from new
tokenized stock and predictions markets; Solana (SOL) - benefiting from expanding tokenization use
cases, including launch of tokenized stocks on Backed Finance.

Downward trend: Bitcoin (BTC) - affected by phishing scams targeting individuals and corporate
impersonation fraud, resulting in significant financial losses and negative sentiment; Crypto in
general - facing increased volatility and risk-off sentiment due to new tariffs, disappointing jobs
data, and a general shift toward risk aversion.

Contradictory trend: None identified—no cryptocurrency has clear contradictory conclusions in the
provided news.

Conclusion: Bitcoin is experiencing a strong upward trend driven by institutional inflows and
regulatory clarity, while Ethereum and Solana are gaining momentum from growing tokenization
adoption; in contrast, Bitcoin is facing downward pressure from scams and fraud, and the overall
crypto market is experiencing heightened volatility due to macroeconomic factors.
\end{verbatim}
}

\subsubsection{Example of stacking graph and text summaries of summaries}
{\fontsize{9}{11}\selectfont \begin{verbatim}
{
  "upward_trend": [
    {
      "Bitcoin": "Experiencing a strong upward trend due to record-high price peaks, significant
institutional accumulation, positive sentiment, and substantial ETF inflows. Factors include
regulatory clarity and strong inflows from institutional investors."
    },
    {
      "Ethereum": "Rising on the back of growing institutional adoption, tokenized real-world asset
offerings, and positive sentiment from new tokenized stock and predictions markets."
    },
    {
      "Solana": "Benefiting from expanding tokenization use cases, including the launch of tokenized
stocks on Backed Finance, leading to increased institutional interest."
    },
    {
      "Crypto in general": "Supported by positive sentiment, institutional activity, and regulatory
progress such as the Genius Act."
    }
  ],
  "downward_trend": [
    {
      "Bitcoin": "Facing significant downward pressure and negative sentiment due to price drops,
phishing scams targeting individuals, corporate impersonation fraud, and financial losses, despite
institutional and regulatory support."
    },
    {
      "Crypto in general": "Experiencing increased volatility and risk-off sentiment due to new
tariffs, disappointing jobs data, and a general shift toward risk aversion in the broader market."
    }
  ],
  "contradictory_trend": [
    {
      "Bitcoin": "Shows contradictory signals—strong upward momentum from institutional activity and
ETF inflows, but also significant downward price movement and negative sentiment from scams and
fraud, leading to mixed market perceptions."
    }
  ],
  "conclusion": "Bitcoin, Ethereum, and Solana are all experiencing strong upward trends driven by
institutional inflows, regulatory clarity, and tokenization adoption. However, Bitcoin faces
contradictory signals from both positive institutional support and negative impacts from scams and
fraud. The overall crypto market is buoyed by positive sentiment and regulatory progress, but is
also affected by macroeconomic factors leading to increased volatility. No clear strong downward
trends are identified in the provided news."
}
\end{verbatim}
}

\section{Conclusion}
In the work, we consider multilevel multitask analysis of cryptocurrency news using the fine-tuned Mistral 7B large language model with retrieval-augmented generation (RAG). 
For fine-tuning, we used 4-bit quantization with Peft/LoRA approach.  On the first level of analytics, the fine-tuned model generates graph and text  
summaries with sentiment scores as well as JSON representations of summaries. Higher levels perform hierarchical stacking that consolidates sets of graph-based and text-based summaries as well as summaries of summaries into comprehensive reports.  
 The use of the summarizing of the list of summaries makes it possible to receive a complex optimized summary of several news where similarity and diversity in cryptocurrency trends can be highlighted. 
Representation cryptocurrency news as knowledge graph can essentially eliminate problems with large language model hallucinations. Using two types of news summaries - knowledge graph and text summaries provides a diversified representation of crypocurrency news. This approach makes it possible  to receive a comprehensive summary on the stacking level. 
The  sentiments signal scores generated by fine-tuned Mistral 7B model can be used in predictive models as features.   
In the test results, some inaccuracy can appear. To improve the LLM performance, one needs a more precisely created training dataset and find optimized parameters for fine-tuning.  
Graph and text summaries can have different sentiment scores for the same news. On the stacking level, the model 
generates an optimized score taking into account the scores in the list of summaries.  
Hierarchical multilevel stacking approaches of news with graph and text summary representation make it possible to receive a diversified summary and highlight strong and contradictory cryptocurrency trends which are considered in different news. That is important for the decision-making on the expert level, allows one to make risk assessment and eliminate LLM hallucinations in news summarization.
 The combined use of graph and text outputs provides complementary views of the same corpus, yielding diversified and more reliable analytics. 
The obtained results demonstrate that the use of fine-tuned Mistral 7B LLM  models for cryptocurrency news analysis  can give informative qualitative and quantitative analytics, providing important insights. 
\section{Disclaimer} 
The approach, ideas, and results shared in this study are for academic purposes only and are not intended to inform real-world conclusions or recommendations.

\bibliographystyle{unsrt}
\bibliography{article.bib}
\FloatBarrier
\newpage
\appendix
\section{Appendix}
\subsection{Python Code for Testing Fine-Tuned Mistral 7B LLM}
{\fontsize{9}{11}\selectfont \begin{verbatim}
from transformers import AutoModelForCausalLM, AutoTokenizer
from peft import PeftModel
from huggingface_hub import login

#Login to Huggingface to load Mistral LLM
login("Huggingface access token")

model_id = "mistralai/Mistral-7B-Instruct-v0.1"
peft_model_name="bpavlsh/Mistral-crypto-news"

#Choose prompt query 
prompt_query_1="Generate a knowledge graph from cryptocurrency news:"
prompt_query_2="Generate summaries of cryptocurrency news and detect sentiment signals:"
prompt_query_3="Create a JSON representation of the summary of cryptocurrency news:"

tokenizer = AutoTokenizer.from_pretrained(model_id)
base_model = AutoModelForCausalLM.from_pretrained( model_id, load_in_4bit=True,
                                  device_map="auto", torch_dtype="auto")
model = PeftModel.from_pretrained(base_model, peft_model_name)

text=""" Cryptocurriency news text from 1Kb to 10Kb """

prompt = f"""<s>[INST] <<SYS>>
You are an expert in analyzing cryptocurrency news.
<</SYS>>

{prompt_query_1}
{text} [/INST]"""

inputs = tokenizer(prompt, return_tensors="pt").to("cuda")
output = model.generate(**inputs, max_new_tokens=1500)
output_result=tokenizer.decode(output[0], skip_special_tokens=True)
result=output_result.split('[/INST]')[1]
print(f"\n{result}")

\end{verbatim}
}

\subsection{Examples of Graph and Text Summaries Generated by Fine-Tuned Mistral 7B LLM} 
{\fontsize{9}{11}\selectfont \begin{verbatim}

Graph summary:

Coinbase — partnered_with — JPMorgan
Coinbase — listed_on — S&P 500 index
Coinbase — has_price_change — +50%
Coinbase — has_sentiment_signal — 8

JPMorgan — partnered_with — Coinbase
JPMorgan — involved_in — crypto-related products
JPMorgan — has_sentiment_signal — 3

Crypto in general — has_trend — upward trend
Crypto in general — has_sentiment_signal — 7

USDC — involved_in — credit card purchase link
USDC — has_sentiment_signal — 6

Stablecoins — has_trend — upward trend
Stablecoins — has_sentiment_signal — 7

Graph summary in JSON format:

 [
  {"Entity 1":"Coinbase","Relationship":"partnered_with","Entity 2":"JPMorgan"},
  {"Entity 1":"Coinbase","Relationship":"listed_on","Entity 2":"S&P 500 index"},
  {"Entity 1":"Coinbase","Relationship":"has_price_change","Entity 2":"+50%"},
  {"Entity 1":"Coinbase","Relationship":"has_sentiment_signal","Entity 2":"8"},
  {"Entity 1":"JPMorgan","Relationship":"partnered_with","Entity 2":"Coinbase"},
  {"Entity 1":"JPMorgan","Relationship":"involved_in","Entity 2":"crypto-related products"},
  {"Entity 1":"JPMorgan","Relationship":"has_sentiment_signal","Entity 2":"3"},
  {"Entity 1":"Crypto in general","Relationship":"has_trend","Entity 2":"upward trend"},
  {"Entity 1":"Crypto in general","Relationship":"has_sentiment_signal","Entity 2":"7"},
  {"Entity 1":"USDC","Relationship":"involved_in","Entity 2":"credit card purchase link"},
  {"Entity 1":"USDC","Relationship":"has_sentiment_signal","Entity 2":"6"},
  {"Entity 1":"Stablecoins","Relationship":"has_trend","Entity 2":"upward trend"},
  {"Entity 1":"Stablecoins","Relationship":"has_sentiment_signal","Entity 2":"7"}
]


Text summary:

 **Summary:**
JPMorgan Chase has partnered with Coinbase to enable Chase credit card users to purchase
cryptocurrency directly on the Coinbase platform starting in fall 2025, signaling further mainstream
adoption of cryptocurrencies. Coinbase shares surged following the news, reflecting growing investor
optimism about the integration of traditional banking with digital assets. The move comes as
regulatory clarity and stablecoin adoption accelerate, with Coinbase and other crypto-related
companies seeing strong market performance.

---

**Crypto Descriptions and Sentiment Scores:**

- **Crypto in general:**
  The partnership and regulatory advancements indicate growing acceptance and maturity of the
cryptocurrency market, with increasing mainstream adoption and use cases.
  **Sentiment signal: +8**

- **Coinbase (COIN):**
  Coinbase shares jumped over 3% after the announcement of JPMorgan’s partnership, reflecting strong
investor sentiment about the exchange’s growing legitimacy and expanding user base.
  **Sentiment signal: +9**

- **USDC (USD Coin):**
  USDC, a stablecoin pegged to the US dollar, will be available for direct redemption via Chase
credit card purchases on Coinbase, likely boosting its utility and adoption.
  **Sentiment signal: +7**

- **JPMorgan Chase (JPM):**
  JPMorgan is taking a significant step into the crypto space by enabling crypto purchases for its
customers, signaling a shift in traditional banking attitudes toward digital assets.
  **Sentiment signal: +6**

---

**Trading Signals:**
  The partnership between JPMorgan and Coinbase is a strong bullish signal for the overall
cryptocurrency market, particularly for crypto exchanges and stablecoins.


Text summary in JSON format:
 
 [{"Cryptocurrency":"Crypto in general","summary":"The partnership and regulatory advancements
indicate growing acceptance and maturity of the cryptocurrency market, with increasing mainstream
adoption and use cases.","sentiment":8},{"Cryptocurrency":"Coinbase (COIN)","summary":"Coinbase
shares jumped over 3% after the announcement of JPMorgan’s partnership, reflecting strong investor
sentiment about the exchange’s growing legitimacy and expanding user
base.","sentiment":9},{"Cryptocurrency":"USDC (USD Coin)","summary":"USDC, a stablecoin pegged to
the US dollar, will be available for direct redemption via Chase credit card purchases on Coinbase,
likely boosting its utility and adoption.","sentiment":7},{"Cryptocurrency":"JPMorgan Chase
(JPM)","summary":"JPMorgan is taking a significant step into the crypto space by enabling crypto
purchases for its customers, signaling a shift in traditional banking attitudes toward digital
assets.","sentiment":6}]

------------------------------------------------------------------------------------------------

Graph summary:

 Bitcoin — has_price_change — -2.8%
Bitcoin — has_trend — downward trend
Bitcoin — affected_by — crypto-related bills blocked in House of Representatives
Bitcoin — affected_by — failed crypto legislation
Bitcoin — has_sentiment_signal — -6

Coinbase — has_price_change — -1.5%
Coinbase — has_trend — downward trend
Coinbase — involved_in — crypto trading platform
Coinbase — has_sentiment_signal — -3

Riot Platforms — has_price_change — -3.3%
Riot Platforms — has_trend — downward trend
Riot Platforms — involved_in — bitcoin mining
Riot Platforms — has_sentiment_signal — -4

Mara Holdings — has_price_change — -2.3%
Mara Holdings — has_trend — downward trend
Mara Holdings — involved_in — bitcoin mining
Mara Holdings — has_sentiment_signal — -3

Crypto in general — affected_by — crypto-related bills blocked in House of Representatives
Crypto in general — affected_by — failed crypto legislation
Crypto in general — has_trend — downward trend
Crypto in general — has_sentiment_signal — -5


Graph summary in JSON format:

 [
  {"Entity 1":"Bitcoin","Relationship":"has_price_change","Entity 2":"-2.8%"},
  {"Entity 1":"Bitcoin","Relationship":"has_trend","Entity 2":"downward trend"},
  {"Entity 1":"Bitcoin","Relationship":"affected_by","Entity 2":"crypto-related bills blocked in
House of Representatives"},
  {"Entity 1":"Bitcoin","Relationship":"affected_by","Entity 2":"failed crypto legislation"},
  {"Entity 1":"Bitcoin","Relationship":"has_sentiment_signal","Entity 2":"-6"},
  {"Entity 1":"Coinbase","Relationship":"has_price_change","Entity 2":"-1.5%"},
  {"Entity 1":"Coinbase","Relationship":"has_trend","Entity 2":"downward trend"},
  {"Entity 1":"Coinbase","Relationship":"involved_in","Entity 2":"crypto trading platform"},
  {"Entity 1":"Coinbase","Relationship":"has_sentiment_signal","Entity 2":"-3"},
  {"Entity 1":"Riot Platforms","Relationship":"has_price_change","Entity 2":"-3.3%"},
  {"Entity 1":"Riot Platforms","Relationship":"has_trend","Entity 2":"downward trend"},
  {"Entity 1":"Riot Platforms","Relationship":"involved_in","Entity 2":"bitcoin mining"},
  {"Entity 1":"Riot Platforms","Relationship":"has_sentiment_signal","Entity 2":"-4"},
  {"Entity 1":"Mara Holdings","Relationship":"has_price_change","Entity 2":"-2.3%"},
  {"Entity 1":"Mara Holdings","Relationship":"has_trend","Entity 2":"downward trend"},
  {"Entity 1":"Mara Holdings","Relationship":"involved_in","Entity 2":"bitcoin mining"},
  {"Entity 1":"Mara Holdings","Relationship":"has_sentiment_signal","Entity 2":"-3"},
  {"Entity 1":"Crypto in general","Relationship":"affected_by","Entity 2":"crypto-related bills
blocked in House of Representatives"},
  {"Entity 1":"Crypto in general","Relationship":"affected_by","Entity 2":"failed crypto
legislation"},
  {"Entity 1":"Crypto in general","Relationship":"has_trend","Entity 2":"downward trend"},
  {"Entity 1":"Crypto in general","Relationship":"has_sentiment_signal","Entity 2":"-5"}
]


Text summary:

 **Summary:**
Bitcoin fell below $117,000 on Tuesday after crypto-related bills in the U.S. House of
Representatives were blocked, ending hopes for immediate crypto regulation. Bitcoin miners Riot
Platforms and Mara Holdings dropped sharply, while Coinbase also declined, reflecting broader market
caution.

---

**Crypto Descriptions and Sentiment:**

- **Bitcoin (BTC):**
  Bitcoin's price dropped below $117,000 following the failure of crypto regulation bills and a
broader risk-off sentiment.
  **Sentiment Signal:** -6

- **Riot Platforms (RIOT):**
  Riot Platforms, a Bitcoin mining company, saw its stock decline by 3.3% amid the market pullback.
  **Sentiment Signal:** -5

- **Mara Holdings (MHRB):**
  Mara Holdings, another Bitcoin mining firm, also experienced a 2.3% drop in its share price.
  **Sentiment Signal:** -4

- **Coinbase (COIN):**
  Coinbase, a leading cryptocurrency exchange, lost 1.5% as investor sentiment soured after setbacks
in crypto legislation.
  **Sentiment Signal:** -4

- **Crypto in general:**
  The broader crypto market faced pressure as optimism over new U.S. crypto regulations faded after
bills were blocked in Congress.
  **Sentiment Signal:** -5

---

**Note:**
No other specific cryptocurrencies were mentioned in the text.


Text summary in JSON format:

 [
  {
    "Cryptocurrency": "Bitcoin",
    "summary": "Bitcoin's price dropped below $117,000 following the failure of crypto regulation
bills and a broader risk-off sentiment.",
    "sentiment": -6
  },
  {
    "Cryptocurrency": "Riot Platforms",
    "summary": "Riot Platforms, a Bitcoin mining company, saw its stock decline by 3.3% amid the
market pullback.",
    "sentiment": -5
  },
  {
    "Cryptocurrency": "Mara Holdings",
    "summary": "Mara Holdings, another Bitcoin mining firm, also experienced a 2.3% drop in its
share price.",
    "sentiment": -4
  },
  {
    "Cryptocurrency": "Coinbase",
    "summary": "Coinbase, a leading cryptocurrency exchange, lost 1.5% as investor sentiment soured
after setbacks in crypto legislation.",
    "sentiment": -4
  },
  {
    "Cryptocurrency": "Crypto in general",
    "summary": "The broader crypto market faced pressure as optimism over new U.S. crypto
regulations faded after bills were blocked in Congress.",
    "sentiment": -5
  }
]

------------------------------------------------------------------------------------------------

Graph summary:

 Crypto regulation bills — failed_to_advance — House of Representatives
Trump — supported_by — Crypto regulation bills
Crypto regulation bills — regulated_by — House of Representatives
Crypto regulation bills — affected_by — Rules of debate vote
Crypto regulation bills — failed — Rules of debate vote
Republican skeptics — blocked — Crypto regulation bills
Crypto regulation bills — has_sentiment_signal — -7

Stablecoin — regulated_by — GENIUS Act
GENIUS Act — failed — Rules of debate vote
GENIUS Act — has_sentiment_signal — -6

CLARITY Act — failed — Rules of debate vote
CLARITY Act — has_sentiment_signal — -6

Federal Reserve — target_of — Crypto regulation bills
Federal Reserve — regulated_by — Crypto regulation bills

Circle — affected_by — Failed Crypto regulation bills
Circle — has_price_change — -7%

Coinbase — affected_by — Failed Crypto regulation bills
Coinbase — has_price_change — -4%

MARA Holdings — affected_by — Failed Crypto regulation bills
MARA Holdings — has_price_change — -2%

Crypto in general — has_sentiment_signal — -7


Graph summary in JSON format:

 [
  {"Entity 1":"Crypto regulation bills","Relationship":"failed_to_advance","Entity 2":"House of
Representatives"},
  {"Entity 1":"Trump","Relationship":"supported_by","Entity 2":"Crypto regulation bills"},
  {"Entity 1":"Crypto regulation bills","Relationship":"regulated_by","Entity 2":"House of
Representatives"},
  {"Entity 1":"Crypto regulation bills","Relationship":"affected_by","Entity 2":"Rules of debate
vote"},
  {"Entity 1":"Crypto regulation bills","Relationship":"failed","Entity 2":"Rules of debate vote"},
  {"Entity 1":"Republican skeptics","Relationship":"blocked","Entity 2":"Crypto regulation bills"},
  {"Entity 1":"Crypto regulation bills","Relationship":"has_sentiment_signal","Entity 2":"-7"},
  {"Entity 1":"Stablecoin","Relationship":"regulated_by","Entity 2":"GENIUS Act"},
  {"Entity 1":"GENIUS Act","Relationship":"failed","Entity 2":"Rules of debate vote"},
  {"Entity 1":"GENIUS Act","Relationship":"has_sentiment_signal","Entity 2":"-6"},
  {"Entity 1":"CLARITY Act","Relationship":"failed","Entity 2":"Rules of debate vote"},
  {"Entity 1":"CLARITY Act","Relationship":"has_sentiment_signal","Entity 2":"-6"},
  {"Entity 1":"Federal Reserve","Relationship":"target_of","Entity 2":"Crypto regulation bills"},
  {"Entity 1":"Federal Reserve","Relationship":"regulated_by","Entity 2":"Crypto regulation bills"},
  {"Entity 1":"Crypto in general","Relationship":"has_sentiment_signal","Entity 2":"-7"},
  {"Entity 1":"Circle","Relationship":"affected_by","Entity 2":"Failed Crypto regulation bills"},
  {"Entity 1":"Circle","Relationship":"has_price_change","Entity 2":"-7%"},
  {"Entity 1":"Coinbase","Relationship":"affected_by","Entity 2":"Failed Crypto regulation bills"},
  {"Entity 1":"Coinbase","Relationship":"has_price_change","Entity 2":"-4%"},
  {"Entity 1":"MARA Holdings","Relationship":"affected_by","Entity 2":"Failed Crypto regulation
bills"},
  {"Entity 1":"MARA Holdings","Relationship":"has_price_change","Entity 2":"-2%"}
]


Text summary:

 **Summary:**
House Republicans who blocked several crypto regulation bills from advancing reversed course and
voted in favor of the measures after a White House meeting with President Trump, following a brief
period of uncertainty. The bills include the GENIUS Act to regulate stablecoins and the CLARITY Act
to establish rules for digital assets, and their failure to pass earlier was a significant setback
for the crypto industry.

---

**Crypto in general:**
The failed procedural vote for crypto regulation bills caused a temporary setback and a drop in
crypto-linked stocks, but a late-night vote reversal following a Trump-led Republican conference
meeting is seen as a positive sign for the industry.
**Sentiment signal:** +3

**Circle (CRCL):**
Shares of Circle, a stablecoin issuer, dropped more than 7% following the failed vote, reflecting
market uncertainty but later rebounded after the bill's passage.
**Sentiment signal:** +2

**Coinbase (COIN):**
Coinbase, a major crypto services provider, saw its stock fall more than 4% after the initial
setback, but also recovered some losses after the bill's eventual passage.
**Sentiment signal:** +2

**MARA Holdings (MARA):**
MARA Holdings, a company involved in crypto mining infrastructure, experienced a similar drop in
share price of about 2% following the initial vote, but also saw a partial recovery.
**Sentiment signal:** +2

**Stablecoins (General Crypto Asset Class):**
Stablecoins are at the center of the legislative debate, with the GENIUS Act aiming to provide
regulatory clarity, which could enhance legitimacy and adoption for the asset class.
**Sentiment signal:** +3

**Bitcoin (BTC):**
Bitcoin experienced a price dip in line with the broader crypto-linked stock market following the
failed vote, but later stabilized as the outcome became clearer.
**Sentiment signal:** 0

**Crypto Regulation Bills (specific legislation):**
Several bills aimed at regulating digital assets and stablecoins, including the GENIUS Act and
CLARITY Act, advanced after a White House meeting, signaling renewed legislative momentum.
**Sentiment signal:** +4


Text summary in JSON format:

 [
  {
    "Cryptocurrency": "Crypto in general",
    "summary": "The failed procedural vote for crypto regulation bills caused a temporary setback
and a drop in crypto-linked stocks, but a late-night vote reversal following a Trump-led Republican
conference meeting is seen as a positive sign for the industry.",
    "sentiment": 3
  },
  {
    "Cryptocurrency": "Circle (CRCL)",
    "summary": "Shares of Circle, a stablecoin issuer, dropped more than 7% following the failed
vote, reflecting market uncertainty but later rebounded after the bill's passage.",
    "sentiment": 2
  },
  {
    "Cryptocurrency": "Coinbase (COIN)",
    "summary": "Coinbase, a major crypto services provider, saw its stock fall more than 4% after
the initial setback, but also recovered some losses.",
    "sentiment": 2
  },
  {
    "Cryptocurrency": "MARA Holdings (MARA)",
    "summary": "MARA Holdings, a company involved in crypto mining infrastructure, experienced a
similar drop in share price of about 2% following the initial vote, but also saw a partial
recovery.",
    "sentiment": 2
  },
  {
    "Cryptocurrency": "Stablecoins (General Crypto Asset Class)",
    "summary": "Stablecoins are at the center of the legislative debate, with the GENIUS Act aiming
to provide regulatory clarity, which could enhance legitimacy and adoption for the asset class.",
    "sentiment": 3
  },
  {
    "Cryptocurrency": "Bitcoin (BTC)",
    "summary": "Bitcoin experienced a price dip in line with the broader crypto-linked stock market
following the failed vote, but later stabilized as the outcome became clearer.",
    "sentiment": 0
  },
  {
    "Cryptocurrency": "Crypto Regulation Bills (specific legislation)",
    "summary": "Several bills aimed at regulating digital assets and stablecoins, including the
GENIUS Act and CLARITY Act, advanced after a White House meeting, signaling renewed legislative
momentum.",
    "sentiment": 4
  }
]

\end{verbatim}
}

\subsection{Examples of Stacking Summaries of Graph/Text Summaries Lists} 
{\fontsize{9}{11}\selectfont \begin{verbatim}

Stacking summary of list of graph summaries:

 Upward trend: Bitcoin (price surged past $113,000, driven by BTC-ETF inflows, public company
purchases, and strong market sentiment), Ethereum (price rose with altcoin rally, supported by
positive sentiment and CoinGecko data), Dogecoin (gains driven by altcoin rally and renewed investor
interest), XRP (gains driven by altcoin rally and positive sentiment), Litecoin (benefited from
broader crypto market momentum and positive sentiment), Stablecoins (gains driven by upward trend,
positive sentiment, and new use cases like credit card purchases), Crypto in general (overall upward
trend, positive sentiment, and increased institutional activity).

Downward trend: None identified with strong downward trend in the provided news.

Contradictory trend: Bitcoin (price has strong upward trend and positive sentiment, but is affected
by significant liquidations and short-term volatility, with sentiment signals ranging from 8 to -6
in different news).

Conclusion: Bitcoin is experiencing a strong upward trend driven by ETF inflows, public company
purchases, and overall market optimism, but faces contradictory signals due to volatility and
liquidations; Ethereum, Dogecoin, XRP, Litecoin, stablecoins, and the overall crypto market are also
in an upward trend supported by positive sentiment and market activity; no clear strong downward
trends are identified in the current news.


Stacking summary of list of text summaries:

Upward trend: Bitcoin (BTC) – driven by institutional inflows, regulatory progress, and strong
retail demand; Ethereum (ETH) – supported by positive market sentiment, regulatory optimism, and
altcoin rally; Dogecoin (DOGE) – benefiting from retail speculation and strong trading volume;
Coinbase (COIN) – experiencing gains from mainstream banking partnerships and tokenization plans;
USDC (USD Coin) – boosted by direct credit card purchase integration with major banks.

Downward trend: None identified with strong negative sentiment in the provided news.

Contradictory trend: Bitcoin (BTC) – shows both strong price gains and declines due to regulatory
setbacks and broader market risk-off sentiment.

Conclusion: Bitcoin is experiencing a strong upward trend due to institutional support and retail
demand, but faces contradictory pressure from regulatory setbacks and market volatility; Ethereum’s
positive momentum is shaped by regulatory optimism and altcoin market strength; Dogecoin’s rally is
driven by retail speculation and high trading volumes; Coinbase is surging on the back of mainstream
banking partnerships and tokenization ambitions; USDC is gaining from increased utility and direct
purchase integration.


Stacking summary of summary of graph summary list and summary of text summary list:

 {
  "upward_trend": [
    {"Bitcoin": "Strong upward trend driven by ETF inflows, public and institutional demand,
positive market sentiment, and retail participation, despite some volatility."},
    {"Ethereum": "Rising with the broader altcoin rally, supported by positive sentiment, regulatory
optimism, and CoinGecko data."},
    {"Dogecoin": "Gains fueled by retail speculation, strong trading volumes, and renewed investor
interest."},
    {"XRP": "Positive momentum from altcoin rally and investor sentiment."},
    {"Litecoin": "Benefiting from broader crypto market momentum and positive market outlook."},
    {"Stablecoins": "Upward trend due to new use cases, such as credit card purchases, and positive
sentiment."},
    {"Coinbase": "Gains from mainstream banking partnerships and tokenization plans."},
    {"USDC": "Boosted by direct credit card purchase integration with major banks, increasing
utility."},
    {"Crypto in general": "Overall upward trend with positive sentiment and increased institutional
activity."}
  ],
  "downward_trend": [],
  "contradictory_trend": [
    {"Bitcoin": "Despite strong upward momentum from institutional and retail inflows, faces
volatility and short-term liquidations, resulting in contradictory sentiment signals."}
  ],
  "conclusion": "The cryptocurrency market is experiencing a broad upward trend, led by Bitcoin,
Ethereum, and other major altcoins due to positive sentiment, institutional support, and regulatory
optimism. Bitcoin, however, shows both strong upward momentum and contradictory volatility due to
regulatory setbacks and market risk-off sentiment. No strong downward trends are identified in the
current news, with stablecoins, Coinbase, and USDC also benefiting from increased utility and
partnerships."
}

------------------------------------------------------------------------------------------------

Stacking summary of list of graph summaries:

 Upward trend: Bitcoin (BTC) - driven by ETF inflows, public company purchases, and strong market
sentiment; Ethereum (ETH) - supported by altcoin rally and positive sentiment; Solana (SOL) -
benefiting from altcoin momentum and positive sentiment; Dogecoin (DOGE) - fueled by renewed
investor interest and altcoin rally; XRP (XRP) - benefiting from broader altcoin rally and positive
sentiment; Litecoin (LTC) - supported by altcoin momentum and positive sentiment; Coinbase (COIN) -
benefiting from crypto stock rally and positive sentiment; Stablecoin legislation - gaining momentum
due to legislative progress.

Downward trend: Crypto in general - faced with stalled U.S. crypto regulation bills, volatility in
crypto stocks, and negative sentiment; MicroStrategy (MSTR) - experienced a slight decline in stock
price after a recent Bitcoin purchase, despite ongoing BTC accumulation; Crypto regulation bills -
hampered by House Republican infighting and last-minute changes.

Contradictory trend: Bitcoin (BTC) - while BTC is experiencing strong upward momentum from ETF
inflows and public purchases, it also faces negative sentiment due to stalled U.S. regulation and
volatility in crypto stocks.

Conclusion: Bitcoin is in a strong upward trend due to ETF inflows and public company purchases, but
faces contradictory signals from regulatory uncertainty and market volatility; Ethereum, Solana,
Dogecoin, XRP, Litecoin, and Coinbase are all riding the altcoin rally and positive sentiment;
Stablecoin legislation shows progress, while the overall crypto market and some stocks are impacted
by regulatory setbacks and volatility.


Stacking summary  of list of text summaries:

 Upward trend: Bitcoin (BTC) - driven by institutional inflows, regulatory progress, and strong
retail demand; Ethereum (ETH) - supported by tokenization of real-world assets, institutional
adoption, and new product launches; Solana (SOL) - benefiting from tokenization of stocks and
growing role in RWA; Stablecoins (general) - boosted by new U.S. legislation enabling dollar-backed
tokens and strengthening regulatory frameworks.

Downward trend: None identified.

Contradictory trend: None identified.

Conclusion: Bitcoin is experiencing a strong upward trend due to robust institutional and retail
demand, positive regulatory developments, and inflows into ETFs; Ethereum is rising on the back of
institutional adoption, tokenization of real-world assets, and new product launches; Solana is
gaining momentum from its role in tokenized stocks and expanding RWA exposure; overall, the crypto
market is seeing positive momentum fueled by regulatory clarity, institutional participation, and
technological innovation.


Stacking summary of summary of graph summary list and summary of text summary list:

 {
  "upward_trend": [
    {"Bitcoin (BTC)": "Experiencing strong upward momentum driven by ETF inflows, public and
institutional purchases, robust retail demand, and positive regulatory progress."},
    {"Ethereum (ETH)": "Supported by altcoin rally, institutional adoption, tokenization of real-
world assets, new product launches, and positive sentiment."},
    {"Solana (SOL)": "Benefiting from altcoin momentum, tokenization of stocks (especially RWA), and
positive sentiment."},
    {"Dogecoin (DOGE)": "Fueled by renewed investor interest and the broader altcoin rally."},
    {"XRP (XRP)": "Benefiting from the broader altcoin rally and positive sentiment."},
    {"Litecoin (LTC)": "Supported by altcoin momentum and positive sentiment."},
    {"Coinbase (COIN)": "Benefiting from the crypto stock rally and positive sentiment."},
    {"Stablecoins (general)": "Boosted by new U.S. legislation enabling dollar-backed tokens and
strengthening regulatory frameworks."}
  ],
  "downward_trend": [
    {"Crypto in general": "Facing challenges from stalled U.S. crypto regulation bills, volatility
in crypto stocks, and negative sentiment."},
    {"MicroStrategy (MSTR)": "Experienced a slight decline in stock price after a recent Bitcoin
purchase, despite ongoing BTC accumulation."},
    {"Crypto regulation bills": "Hindered by House Republican infighting and last-minute changes,
slowing progress in the U.S. regulatory landscape."}
  ],
  "contradictory_trend": [
    {"Bitcoin (BTC)": "While BTC is experiencing strong upward momentum from ETF inflows and
public/institutional purchases, it also faces negative sentiment due to regulatory uncertainty and
market volatility."}
  ],
  "conclusion": "The overall cryptocurrency market is seeing a strong upward trend, led by Bitcoin,
Ethereum, Solana, and other major altcoins due to institutional inflows, positive regulatory
developments, and technological innovation. However, regulatory setbacks and market volatility
present some downside risks, particularly for Bitcoin and the broader market. Bitcoin stands out for
its contradictory signals, as it benefits from both strong fundamentals and regulatory
uncertainties, reflecting the current dynamic landscape."
}

------------------------------------------------------------------------------------------------

Stacking summary of list of graph summaries:

Upward trend:
Bitcoin (BTC): Spot Bitcoin ETFs have experienced 12 days of sustained inflows, indicating strong
institutional adoption and positive sentiment;
Ethereum (ETH): Ethereum ETFs have seen significant inflows and a large single-day inflow, with
positive sentiment and growing institutional interest;
Stablecoins (general): Regulatory clarity and increased mainstream adoption are driving an upward
trend and strong sentiment.

Downward trend:
CoinDCX: Suffered a major $44 million hack, leading to negative sentiment and concerns about
security.

Contradictory trend:
Crypto in general: While overall positive sentiment and regulatory progress are noted, there are
also concerns about cybersecurity threats and potential risks from private corporate currencies.

Conclusion:
Bitcoin is experiencing a strong upward trend driven by surging institutional inflows and positive
sentiment in spot ETFs;
Ethereum is benefiting from significant ETF inflows and growing institutional interest;
Stablecoins are gaining momentum due to regulatory clarity and mainstream adoption;
CoinDCX faces a downward trend after a major hack undermining trust and security;
The broader crypto market shows contradictory signals, with progress in regulation and adoption
tempered by persistent cybersecurity threats and industry debates.


Stacking summary  of list of text summaries:

Upward trend:
Bitcoin (BTC) – driven by institutional inflows into spot ETFs, mainstream banking partnerships
(JPMorgan), and growing legitimacy;
USDC – boosted by direct redemption access via JPMorgan credit card purchases on Coinbase;
Coinbase (COIN) – positive sentiment from mainstream banking partnerships, new product launches, and
expanding crypto offerings.

Downward trend:
Grayscale Bitcoin Trust (GBTC) – experiencing price decline and reduced demand after incentive
program expiration despite increased trading volume;
Crypto in general – facing increased volatility and risk-off sentiment due to macroeconomic factors,
regulatory uncertainty, and disappointing jobs data.

Contradictory trend:
Ethereum (ETH) – shows slight decline in price but maintains overall positive sentiment due to
continued institutional inflows and expanding market structure.

Conclusion:
Bitcoin is surging on the back of institutional adoption and mainstream financial integration; USDC
is gaining traction as a trusted stablecoin with new payment channels; Coinbase is rising on growing
legitimacy and product expansion; GBTC is declining as its incentive program ends despite increased
trading; the broader crypto market faces mixed signals, with volatility and risk aversion from macro
factors offset by positive institutional and regulatory developments.

Summary:
Bitcoin is experiencing strong upward momentum driven by institutional inflows and mainstream
financial integration, while USDC and Coinbase are also benefiting from increased adoption and new
products; Grayscale Bitcoin Trust (GBTC) is declining as its incentive program ends, and the overall
crypto market faces contradictory signals with both volatility and positive institutional
developments.

Upward trend: Bitcoin (BTC), USDC, Coinbase.
Downward trend: Grayscale Bitcoin Trust (GBTC).
Contradictory trend: Ethereum (ETH), Crypto in general.
Conclusion: Bitcoin is surging on institutional inflows and mainstream adoption, USDC and Coinbase
are gaining traction as adoption increases, Grayscale Bitcoin Trust is declining as its incentive
program ends, Ethereum and the broader crypto market show mixed signals with both volatility and
positive institutional developments, highlighting both risks and opportunities for investors.


Stacking summary of summary of graph summary list and summary of text summary list:

 {
  "upward_trend": [
    {
      "Bitcoin": "Experiencing strong upward momentum due to sustained institutional inflows into
spot ETFs, growing legitimacy, mainstream banking partnerships (e.g., JPMorgan), and expanding
market structure."
    },
    {
      "Ethereum": "Benefiting from significant ETF inflows and growing institutional interest, with
continued positive sentiment despite slight price declines."
    },
    {
      "Stablecoins": "Gaining momentum from regulatory clarity, increased mainstream adoption, and
new payment channels like direct redemption access via JPMorgan credit card purchases on Coinbase."
    },
    {
      "Coinbase": "Rising on mainstream banking partnerships, new product launches, expanding crypto
offerings, and increased legitimacy."
    }
  ],
  "downward_trend": [
    {
      "CoinDCX": "Suffering a major $44 million hack, leading to negative sentiment, trust issues,
and concerns about security."
    },
    {
      "Grayscale Bitcoin Trust (GBTC)": "Declining due to the expiration of incentive programs,
reduced demand, and price declines, despite increased trading volume."
    }
  ],
  "contradictory_trend": [
    {
      "Crypto in general": "Shows both positive signals from regulatory progress and mainstream
adoption, but is tempered by persistent cybersecurity threats, regulatory uncertainty, and
macroeconomic volatility."
    },
    {
      "Ethereum": "Despite slight price decline, overall positive sentiment and continued
institutional inflows provide a mixed outlook."
    }
  ],
  "conclusion": "Bitcoin is leading the upward trend on institutional adoption and mainstream
financial integration, while Ethereum and stablecoins also benefit from positive sentiment and
growing legitimacy. CoinDCX and GBTC face downward pressure from hacks and expired incentives. The
broader crypto market exhibits contradictory signals, with regulatory and adoption progress offset
by cybersecurity risks and macroeconomic volatility."
  }

\end{verbatim}
}

\end{document}